\documentclass[conference]{IEEEtran}
\IEEEoverridecommandlockouts
\usepackage{cite}
\usepackage{placeins}
\usepackage{amsmath,amssymb,amsfonts}
\usepackage{algorithmic}
\usepackage{graphicx}
\usepackage{textcomp}
\usepackage{xcolor}
\def\BibTeX{{\rm B\kern-.05em{\sc i\kern-.025em b}\kern-.08em
   T\kern-.1667em\lower.7ex\hbox{E}\kern-.125emX}}
\begin{document}

\title{MambaLoc: Lightweight Indoor Localization Using Cross-Modal Knowledge Distillation}

\author{
 \IEEEauthorblockN{Mohab Bahnassy\textbf{$^*$}}
 \IEEEauthorblockA{\textit{AUC, Egypt}\\
 mohabbahnassy@aucegypt.edu}
 \and
 \IEEEauthorblockN{Omar Saqr\textbf{$^*$}}
 \IEEEauthorblockA{\textit{AUC, Egypt}\\
 omar\_saqr@aucegypt.edu}
 \and
 \IEEEauthorblockN{Hamada Rizk}
 \IEEEauthorblockA{\textit{The University of Osaka, Japan}\\
 \textit{RIKEN-CCS, Japan}\\
 hamada\_rizk@ist.osaka-u.ac.jp}
 \and
 \IEEEauthorblockN{Moustafa Youssef}
 \IEEEauthorblockA{\textit{AUC, Egypt}\\
 moustafa-youssef@aucegypt.edu}
 \thanks{\textbf{$^*$Both authors contributed equally to this research.}}
}
\IEEEaftertitletext{\vspace{-30pt}}
\maketitle

\begin{abstract}
Accurate indoor localization on resource-constrained edge devices remains challenging due to the tension between modeling fidelity and computational efficiency. Wi-Fi Channel State Information (CSI) enables low-cost deployment using existing Wi-Fi infrastructure but is heavily distorted by multipath propagation. Transformer-based solutions mitigate these effects but their quadratic complexity limits real-time edge deployment. Ultra-Wideband (UWB) provides geometrically precise ranging yet requires dedicated infrastructure. We present \textbf{MambaLoc}, a cross-modal distillation framework that improves the accuracy--efficiency balance in CSI-based localization. MambaLoc uses a linear-time Selective State Space Model (Mamba) as a lightweight CSI student, guided during training by a UWB-Transformer teacher. A \emph{Feature Alignment Module} bridges the representational gap between heterogeneous RF modalities, allowing the CSI model to inherit UWB-derived geometric structure without UWB hardware at inference. A probabilistic Gaussian Mixture regression head models multi-modal spatial ambiguity from multipath effects, avoiding the physically implausible averaging of deterministic regression. Evaluations on a real-world dataset show MambaLoc achieves a median localization error of \textbf{0.230\,m}, reducing error by \textbf{22.2\%} over a standalone Mamba baseline, while requiring \textbf{65.1\%} fewer FLOPs and \textbf{58\%} lower inference latency than Transformer-based approaches.
\end{abstract}


\section{Introduction}
Accurate indoor localization is critical for autonomous systems in GPS-denied environments, including warehouse robotics, drone navigation, and wearable IoT devices. While GNSS provides robust outdoor positioning, signals degrade indoors due to attenuation and multipath propagation \cite{zafari2019survey}. As computation shifts to the edge, localization must deliver both high accuracy and efficient inference under strict hardware constraints.

RF-based approaches leverage existing wireless infrastructure. Wi-Fi CSI provides fine-grained amplitude and phase measurements across multiple subcarriers, implicitly encoding environmental geometry. However, indoor RF propagation is dominated by multipath effects: reflected and diffracted components arrive with different delays and phase shifts, distorting direct range estimates and introducing unreliability \cite{molisch2012wireless}. Localization from CSI therefore requires models that integrate information across packet sequences, separating persistent geometric patterns from short-term noise.

Deep learning has substantially improved CSI-based localization. Transformer architectures model long-range temporal structure via self-attention (e.g., TIPS \cite{zhang2022tips}), but self-attention scales quadratically ($\mathcal{O}(N^2)$), making continuous inference impractical under Size, Weight, and Power (SWaP) constraints.

To address this trade-off, we introduce \textbf{MambaLoc}, built on Selective State Space Models (SSMs). By leveraging the Mamba architecture \cite{gu2024mamba}, our approach achieves linear-time ($\mathcal{O}(N)$) inference with constant memory state. We further adopt a Cross-Modal Knowledge Distillation strategy inspired by the Learning Using Privileged Information (LUPI) paradigm \cite{vapnik2015lupi}: UWB data serves as privileged training context, and a UWB-Transformer teacher guides a lightweight CSI-Mamba student via a Feature Alignment Module. The framework can be interpreted as learning a latent spatial representation of RF propagation, where the teacher provides geometric supervision that guides the CSI model toward physically consistent spatial embeddings. The remainder of this paper is organized as follows: Section~\ref{sec:related} reviews related work; Section~\ref{sec:background} covers theoretical foundations; Sections~\ref{sec:Methodology} and~\ref{sec:training} describe the MambaLoc architecture and distillation curriculum; Section~\ref{sec:eval} presents empirical results.

Our key contributions are: \textbf{(1) Cross-modal RF knowledge distillation}, a framework transferring geometric structure from UWB CIR to WiFi CSI across both sensing modalities (UWB$\rightarrow$CSI) and architectures (Transformer$\rightarrow$Mamba), enabling robust localization without UWB at inference; \textbf{(2) efficient CSI sequence modeling} via Selective State Space Models as an alternative to Transformers ($0.39$ vs.\ $1.12$ GFLOPs); \textbf{(3) a cross-modal feature alignment module} that projects teacher and student representations into a shared latent space, stabilizing distillation and reducing modality mismatch; and \textbf{(4) a probabilistic spatial regression} head using a GMM to model multipath-induced uncertainty.

\section{Related Work}
\label{sec:related}
\subsection{Sensing Modalities for Indoor Localization}

Indoor localization modalities carry distinct trade-offs against edge deployment requirements: SWaP, robustness, and infrastructure cost \cite{zafari2019survey}. Vision SLAM \cite{mur2015orbslam} and LiDAR \cite{zhang2014loam,rizk2023laser} approach centimeter-level precision but demand high-throughput processing, degrade in low-light or texture-poor environments, introduce privacy concerns, and add significant payload \cite{alkendi2021vision}. UWB offers an excellent accuracy--efficiency balance but requires installing and maintaining a purpose-built anchor network whose geometry critically determines coverage quality; NLOS can still cause outliers even in UWB, making large-scale deployment non-trivial \cite{zafari2019survey}. Geometry-based RF methods (ToA, TDoA, AoA) need no dedicated hardware but are acutely sensitive to NLOS conditions \cite{hara2011toa}; high-quality AoA additionally demands antenna arrays and tight clock synchronization. Wi-Fi FTM/RTT and BLE lower the infrastructure bar further~\cite{mohsen2024timesense}, but FTM depends on compatible access points while BLE suffers RSSI instability and requires dense deployment \cite{zafari2019survey}\cite{rizk2022robust}. Sensor fusion mitigates individual weaknesses but multiplies cost, power, and calibration complexity, which is contrary to lightweight edge deployment.

Commodity Wi-Fi CSI occupies a uniquely attractive position: available on already-deployed infrastructure with no additional hardware, it provides per-subcarrier complex channel estimates encoding rich multipath structure far beyond coarse RSSI. The central challenge is extracting reliable spatial features from noisy, high-dimensional CSI, which brings us to deep learning approaches.

\subsection{Deep Learning for CSI Localization}

Classical RF localization relies on geometric properties (ToA, TDoA, AoA) \cite{hara2011toa,alsadoon2020aoa}, often assuming reliable LOS propagation; multipath and NLOS conditions introduce severe biases \cite{fokin2018tdoa}. In contrast, data-driven approaches map CSI fingerprints directly to locations. Early work used MLPs~\cite{wang2017deepfi} and stacked denoising autoencoders with
probabilistic frameworks~\cite{abbas2019wideep}; hybrid CNN-LSTM architectures~\cite{cnn-lstm}
added temporal modeling. Attention-augmented CNNs \cite{zhang2023csi} capture both local and global context. Transformer-based models (TIPS \cite{zhang2022tips}, Swin-Loc \cite{xu2024swinloc}) achieve state-of-the-art accuracy; Transformer-CNN hybrids \cite{uwbtrans} combine global temporal modeling with local spatial extraction for UWB NLOS classification. However, quadratic complexity is prohibitive for edge deployment. Quantization and distillation \cite{foti2025optimising} compress Transformers for microcontrollers, yet even compressed models struggle under tight memory budgets. This motivates Mamba \cite{gu2024mamba}, which offers linear complexity while remaining suitable for edge hardware.

Cross-modal supervision aligns with the LUPI paradigm \cite{vapnik2015lupi}: high-precision sensors guide models that ultimately rely on cheaper sensors. DistilLoc \cite{rizk2024sensors} transfers knowledge from an RTT-based teacher to an RSSI-based student using F-Net, which replaces self-attention with FFT-based mixing ($\mathcal{O}(N\log N)$). MambaLoc goes further with a strictly linear $\mathcal{O}(N)$ Mamba student and cross-modal, cross-architecture supervision, using UWB as a geometric teacher to enforce alignment between the student's CSI latent space and UWB-derived spatial structure that would otherwise be unrecoverable from noisy Wi-Fi data alone.

\section{Background \& Theoretical Foundations}
\label{sec:background}

\subsection{Sequence Modeling with Structured State-Space Models}

Indoor localization using CSI is inherently a sequential inference problem requiring long temporal dependency modeling in high-resolution wireless signals. Structured State-Space Models (SSMs) have re-emerged as a compelling alternative to attention-based architectures. S4 \cite{gu2022s4} and Mamba \cite{gu2024mamba} combine continuous-time dynamical systems with the efficiency of recurrent models. Mamba's selective scan mechanism conditions state transition parameters on the current input $x_t$, enabling adaptive propagation or suppression of information, which is beneficial for discriminating LOS signals from multipath noise. During training, Mamba leverages parallel scan operations; at inference it runs as a recurrent model with constant memory state, processing long sequences in linear time on resource-constrained edge hardware.

\subsection{Learning Using Privileged Information (LUPI)}

We adopt the LUPI paradigm \cite{vapnik2015lupi}, defining the training triplet $(x_{\text{csi}}, x^*_{\text{uwb}}, y_{\text{pos}})$, where $x_{\text{csi}}$ is the deployable modality and $x^*_{\text{uwb}}$ is the privileged modality available only during training. A teacher $f^*:\mathcal{X}^*\rightarrow \mathcal{Y}$ is trained on UWB data to learn a robust geometric mapping; a student $f:\mathcal{X}\rightarrow \mathcal{Y}$ then trains on CSI to mimic the teacher's internal feature representations. By aligning the student's latent space with the teacher's, the student effectively learns to filter CSI multipath effects using heuristics derived from the cleaner UWB domain, all without requiring UWB hardware at inference.

\subsection{Aleatoric Uncertainty in Wireless Localization}

Indoor multipath creates scenarios where different locations produce identical signal fingerprints, an inherent ambiguity termed \textbf{aleatoric uncertainty} \cite{kendall2017uncertainties}. Standard regression models confronted with two equally plausible locations (e.g., the true location and a strong reflection) tend to predict their average, often resulting in impossible locations. MambaLoc addresses this with a \textbf{Gaussian Mixture Model (GMM)} head that predicts a full probability distribution, explicitly capturing multi-modal ambiguity. Rather than maximizing likelihood (which can be unstable), we minimize the distance between ground truth and the distribution's expected value, capturing the uncertainty structure while providing stable gradients \cite{cao2025gmm}.

\begin{figure}[!t]
 \centering
 \includegraphics[width=\columnwidth]{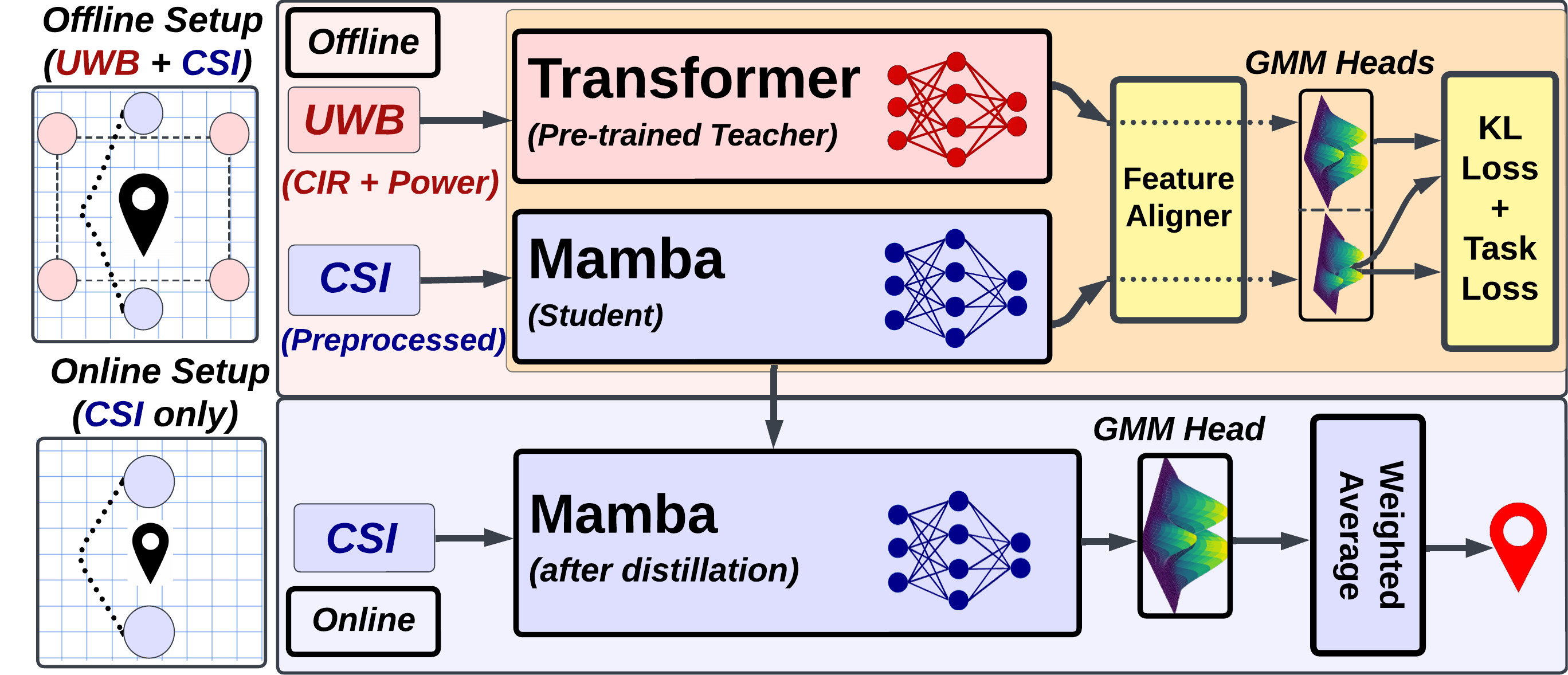}
 \caption{\textbf{MambaLoc system architecture.} The UWB Transformer teacher guides the CSI Mamba student via the Feature Alignment Module during training; at inference, only the Mamba student runs on CSI data.}
 \label{fig:arch}
\end{figure}
\section{Methodology}
\label{sec:Methodology}
\subsection{System Overview}

\textbf{MambaLoc} transfers spatial and geometric knowledge from a high-fidelity \textbf{UWB-based teacher} to a computationally efficient \textbf{CSI-based student} following the LUPI paradigm \cite{vapnik2015lupi}. As shown in Figure~\ref{fig:arch}, the framework operates in two phases:

\begin{itemize}
   \item \textbf{Offline Cross-Modal Distillation:} A UWB-Transformer teacher trained on CIR measurements provides probabilistic supervision and intermediate features. The CSI-Mamba student aligns its latent representations with the teacher through a multi-level distillation pipeline.
   \item \textbf{Online Inference:} After training, the teacher and all UWB components are discarded. The CSI-Mamba student performs real-time localization independently, enabling low-latency inference on edge devices.
\end{itemize}

A key innovation lies in the Feature Alignment Module, which aligns intermediate representations across heterogeneous modalities and architectures rather than distilling only final outputs.

\subsection{Data Representation and Preprocessing}

\subsubsection{UWB Teacher Input}
The teacher processes UWB Channel Impulse Response (CIR) sequences $x_{u} \in \mathbb{R}^{T \times D_u}$, where $T$ denotes temporal snapshots and $D_u$ the feature dimensionality. The input comprises complex CIR magnitude and phase, capturing multipath structure where early peaks correspond to LOS and dominant reflections \cite{jimenez2021decawave}. We augment with normalized first-path power level and total received power level as NLOS discriminators, since signal attenuation patterns differ significantly between LOS and NLOS states. All features are standardized (zero mean, unit variance).

\subsubsection{CSI Student Input}
The student consumes Wi-Fi CSI matrices $x_{c} \in \mathbb{R}^{T \times D_c}$, supporting passive, device-free localization\cite{youssef2007challenges}. CSI is extracted from IEEE 802.11n frames using a $3{\times}3$ MIMO configuration across $N_{sc}{=}30$ subcarriers, yielding 270 complex measurements per packet. Concatenating amplitude and phase gives $D_c = 2 \times 270 = 540$. Raw phase is distorted by sampling frequency offset and packet detection delay; a linear phase sanitization procedure is applied prior to learning \cite{wang2017deepfi}.

\subsection{Network Architectures}

\subsubsection{UWB-Transformer Teacher}
An encoder-only Transformer \cite{zhang2022tips} maps normalized CIR inputs to a latent space of dimension $d_t{=}256$ via a linear embedding, followed by $L{=}6$ encoder layers with $H{=}8$ attention heads. Multi-head self-attention captures global temporal context across variable-length CIR sequences, learning complex multipath and temporal correlations critical for accurate ranging. Its quadratic pairwise interaction ($\mathcal{O}(N^2)$) is acceptable in an offline training role. The output feeds a probabilistic GMM regression head.

\begin{figure}[h]
   \centering
   \includegraphics[width=0.7\linewidth]{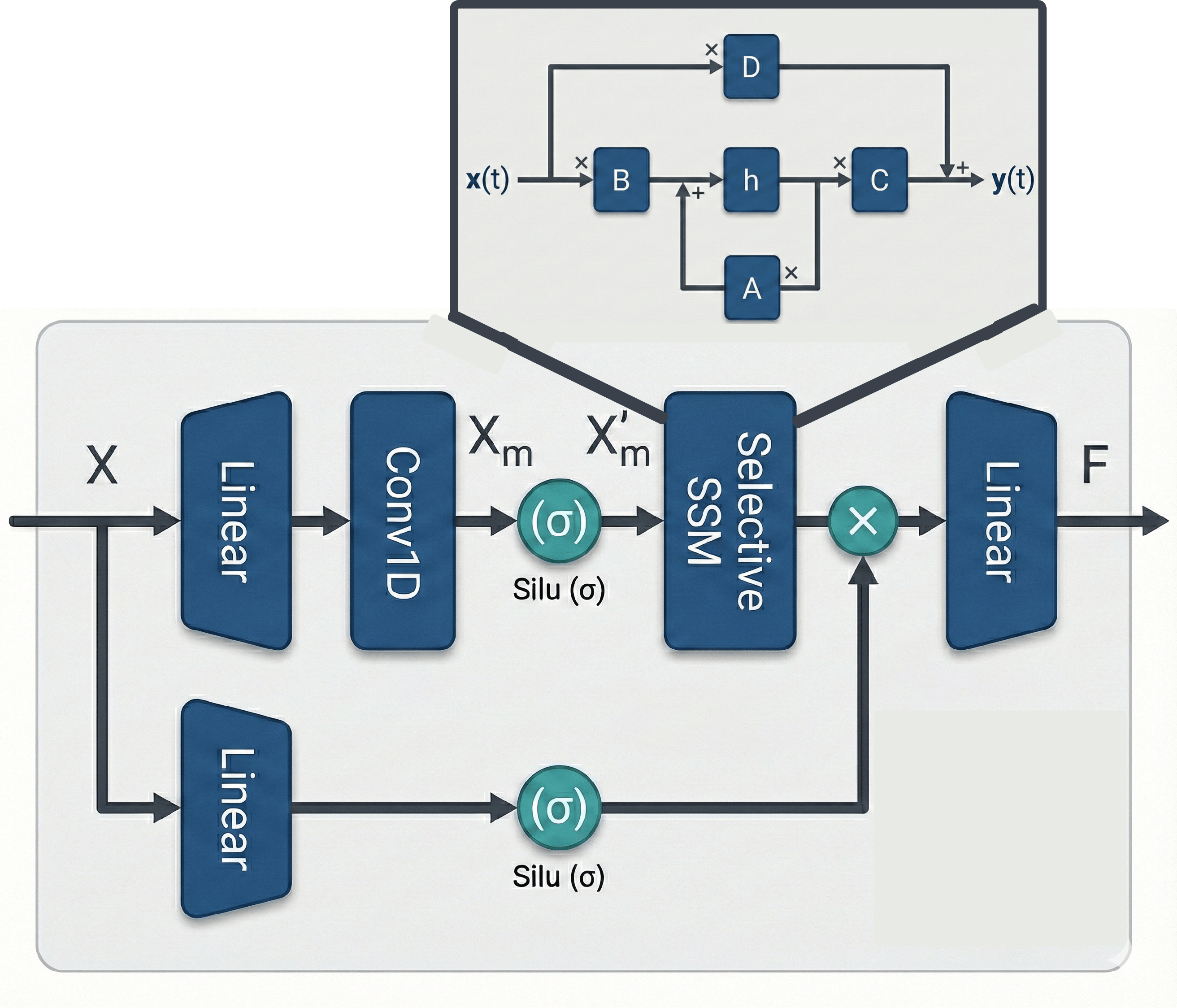}
   \caption{\textbf{Mamba Block Architecture.} Processes CSI features using a 1D convolution and a Selective SSM for efficient, linear-time sequence modeling.}
   \label{fig:mamba}
\end{figure}

\subsubsection{CSI-Mamba Student}
The student uses the Mamba architecture \cite{gu2024mamba} for linear-time complexity. As shown in Figure~\ref{fig:mamba}, each Mamba block processes input $X$ through two parallel branches. The main branch expands features, applies a short \texttt{Conv1D} to capture local temporal dependencies, and then runs the Selective SSM:
\begin{equation}
h_t = \mathbf{A} h_{t-1} + \mathbf{B} x_t, \quad
y_t = \mathbf{C} h_t + \mathbf{D} x_t.
\end{equation}
Unlike fixed LTI systems, parameters $\mathbf{B}$ and $\mathbf{C}$ are computed as functions of $x_t$, allowing the model to dynamically vary its memory, selectively propagating relevant multipath features while filtering transient CSI noise or hardware artifacts. Matrix $\mathbf{D}$ acts as a direct feedthrough (similar to a residual link). The SSM output is gated by the second branch via element-wise multiplication. This design achieves expressive capacity comparable to Transformers \cite{patro2024mamba360} at linear cost.

\subsection{Feature Alignment Module}

\begin{figure}[h]
   \centering
   \includegraphics[width=\linewidth]{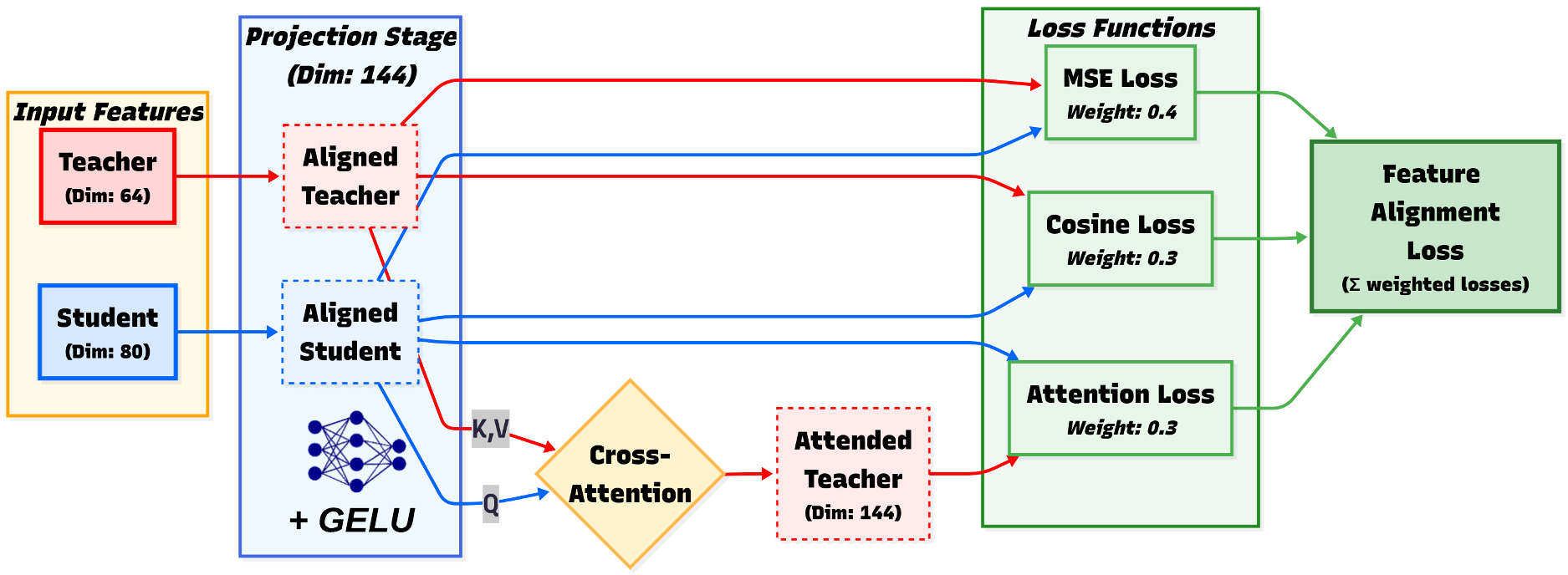}
   \caption{\textbf{Feature Alignment Module architecture.} Projects teacher (UWB) and student (CSI) features into a shared 144-dimensional space using MSE, Cosine Similarity, and Cross-Attention losses.}
   \label{fig:efam}
\end{figure}

Directly distilling final coordinate outputs is insufficient due to the substantial modality gap between UWB and CSI and the scarcity of structural information in a 2D coordinate. Inspired by cross-modal distillation literature \cite{hinton2015,gupta2016crossmodal,rizk2024sensors}, we introduce a \textbf{Feature Alignment Module} (Figure~\ref{fig:efam}).

Let $F_u \in \mathbb{R}^{B \times d_u}$ and $F_c \in \mathbb{R}^{B \times d_c}$ denote teacher and student features. The module performs:
\begin{enumerate}
   \item \textbf{Non-Linear Projection:} Both feature sets are projected into a shared $d_a$-dimensional alignment space via separate two-layer MLPs with LayerNorm and GELU activations: $F'_u = \phi_u(F_u)$, $F'_c = \phi_c(F_c)$.
   \item \textbf{Cross-Modal Attention:} Student features act as queries against teacher keys and values, allowing the CSI student to selectively extract the geometric features most relevant to its current noisy signal:
   \begin{equation}
       \hat{F}_u = \text{Attention}(Q=F'_c,\ K=F'_u,\ V=F'_u).
   \end{equation}
\end{enumerate}

The alignment loss combines direct feature matching, directional consistency, and attention-guided structural alignment:
\begin{equation}
\mathcal{L}_{feat} =
\lambda_{dir} \|F'_u - F'_c\|_2^2 +
\lambda_{cos} \left(1 - \cos(F'_u, F'_c)\right) +
\lambda_{attn} \|\hat{F}_u - F'_c\|_2^2
\end{equation}
where $\lambda_{dir}{=}0.4$, $\lambda_{cos}{=}0.3$, $\lambda_{attn}{=}0.3$ (selected by grid search on the validation set). The first term enforces direct feature matching, the second promotes directional consistency, and the third ensures the student aligns with relevant structural information from the teacher.

\subsection{Probabilistic GMM Regression Head}

Both teacher and student use a $K{=}16$ component Gaussian Mixture head ($K$ selected empirically from $\{4,8,16,32\}$ by lowest validation error). The predicted density is:
\begin{equation}
p(\mathbf{y}|\mathbf{x}) = \sum_{k=1}^{K} \pi_k(\mathbf{x}) \mathcal{N}(\mathbf{y}|\boldsymbol{\mu}_k(\mathbf{x}), \Sigma_k(\mathbf{x})).
\end{equation}
For simplicity, the covariance is treated as a fixed isotropic variance and is not explicitly trained. Rather than maximizing likelihood directly, the primary objective minimizes the error of the expected coordinate estimate $\hat{\mathbf{y}} = \sum_k \pi_k \boldsymbol{\mu}_k$:
\begin{equation}
\mathcal{L}_{task} = \| \hat{\mathbf{y}} - \mathbf{y}_{gt} \|_2^2.
\end{equation}
During distillation, KL divergence between teacher and student mixture weights encourages the student to inherit the teacher's uncertainty structure: $\mathcal{L}_{KD} = \text{KL}(\boldsymbol{\pi}_T \,\|\, \boldsymbol{\pi}_S)$. We distill only the mixture weights $\boldsymbol{\pi}$, not the full GMM parameters, because matching which regions of the room are spatially probable is the most critical and transferable aspect of ambiguity structure across the modality gap; enforcing agreement on exact Gaussian shapes is a finer-grained detail that CSI and UWB encode at fundamentally different resolutions. Notably, replacing the GMM head with a deterministic regression head caused the distilled student to underperform the standalone CSI-Mamba baseline, confirming the probabilistic formulation is necessary for stable knowledge transfer.

\subsection{Training Objective}

The total loss combines task supervision, feature alignment, and distribution-level distillation with dynamic weights:
\begin{equation}
\mathcal{L}_{total} =
\alpha(t)\, \mathcal{L}_{KD} +
\beta(t)\, \mathcal{L}_{task} +
\gamma(t)\, \mathcal{L}_{feat},
\end{equation}
where $\alpha(t)$, $\beta(t)$, and $\gamma(t)$ are dynamically adjusted via a curriculum schedule described in Section~\ref{sec:training}. After training, only the CSI-Mamba student is retained, achieving Transformer-level accuracy at linear-time inference cost \cite{letaief2022edgeAI}.

\section{Training \& Distillation Strategy}
\label{sec:training}

Training MambaLoc must bridge two simultaneous gaps: (i) a \textbf{modality gap} between UWB CIR and Wi-Fi CSI, and (ii) an \textbf{architectural gap} between the Transformer teacher and the Mamba student. Naive end-to-end distillation leads to unstable optimization. We adopt a three-stage Cross-Modal Distillation curriculum following LUPI principles, where UWB is available only during training and acts as a geometric oracle.

\subsection{Stage I: Teacher Pre-Training}

The UWB-Transformer teacher is trained independently on UWB CIR measurements and ground-truth coordinates using MSE loss, and although optimized for regression, it implicitly learns a probability distribution over the spatial grid via its GMM weights, producing richer supervisory signals than deterministic regression. Teacher parameters are then frozen.

\subsection{Stage II: Cross-Modal Aligner Warming}

Before student distillation, the Feature Alignment Module is pre-trained for a few epochs, with both teacher and student backbones frozen, to map CSI features toward UWB features. This initialization ensures the alignment loss provides meaningful gradients at the start of Stage III, while full training of the module continues in that phase.

\subsection{Stage III: Student Distillation with Adaptive Curriculum}

The CSI-Mamba student trains jointly under three supervision signals.

\subsubsection{Task Supervision}
Direct MSE supervision on the student's own GMM expected output ensures the student remains grounded in real CSI observations rather than over-depending on the teacher, preserving robustness in CSI-only deployment.

\subsubsection{Feature Alignment}
The weighted Euclidean, cosine, and cross-attention loss from Section~\ref{sec:Methodology} acts as a geometric regularizer, encouraging CSI features, which are often dominated by noise, to align with the physically grounded UWB representation.

\subsubsection{Probabilistic Knowledge Distillation}
KL divergence with temperature annealing:
\begin{equation}
\mathcal{L}_{KD} = T(t)^2 \cdot \text{KL}\big(\boldsymbol{\pi}_T(y/T) \,\|\, \boldsymbol{\pi}_S(y/T)\big),
\end{equation}
where $T(t)$ decays linearly from $30.0$ to $10.0$. A high initial temperature softens the teacher's distribution, helping the student learn global spatial geometry; progressive annealing then sharpens predictions toward precise localization.

\subsubsection{Adaptive Weighting}
$\alpha(t)$ and $\beta(t)$ are adjusted based on loss trends: if a component deteriorates, its weight increases. The feature alignment weight $\gamma(t)$ follows a curriculum decay, starting high to guide early representation learning and decreasing as the student develops CSI-based independence. These strategies were found to stabilize training significantly compared to fixed-weight alternatives through extensive experiments.

\section{Experimental Evaluation}
\label{sec:eval}
\subsection{Dataset \& Evaluation Metrics}

We evaluate on \textbf{OPERAnet} \cite{Bocus2022OPERANet}, which to our knowledge is the only publicly available dataset providing synchronized UWB and CSI measurements under a shared ground-truth positioning system, making it the canonical benchmark for this problem. Data was collected in two furnished indoor environments ($5.09\,\mathrm{m}{\times}10.02\,\mathrm{m}$ and $7.20\,\mathrm{m}{\times}7.24\,\mathrm{m}$) over approximately 8 hours, covering six daily activities: walking, sitting, standing, lying down, standing up, and body rotation.

\begin{itemize}
   \item \textbf{CSI:} Intel 5300 NICs ($5$\,GHz), 1600\,Hz packet rate; one transmitter and two receivers in LOS and bi-static configurations; $3{\times}3$ MIMO, 30 subcarriers per receiver.
   \item \textbf{UWB:} Two synchronized passive systems (Decawave EVK1000 and MDEK1001) providing CIR data at ${\approx}1.0016$\,ns temporal resolution.
   \item \textbf{Ground Truth:} High-resolution 2D positions from a dedicated UWB anchor-tag system; all modalities synchronized via NTP ($<20$\,ms).
\end{itemize}

The primary metric is Median Absolute Error, measured as Euclidean distance between predicted and ground-truth locations. Median Absolute Error is robust to outliers and reflects typical localization performance.

\subsection{Comparison with State-of-the-Art}

\begin{figure}[h]
   \centering
   \includegraphics[width=0.6\linewidth]{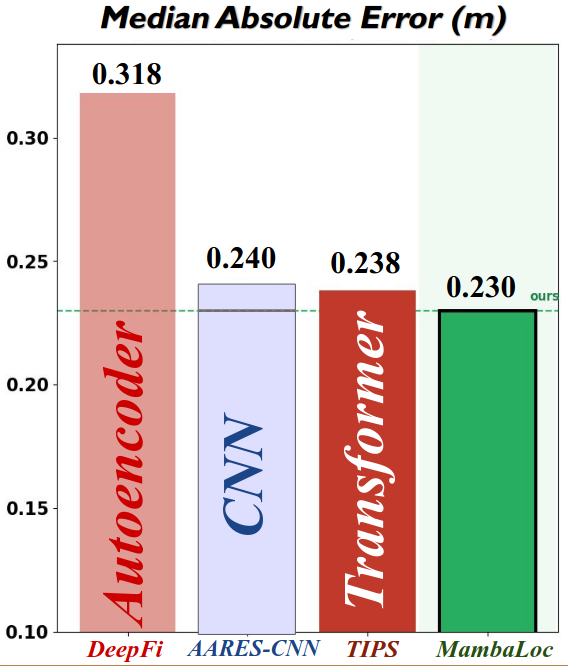}
   \caption{\textbf{Comparison with State-of-the-Art.} MambaLoc achieves the lowest Median AE ($0.230\,\mathrm{m}$), outperforming CNN and Transformer-based baselines.}
   \label{fig:sota_comparison}
\end{figure}

We compare against three representative baselines, all retrained from scratch on OPERAnet for a fair comparison: \textbf{DeepFi}~\cite{wang2017deepfi}, an autoencoder-based compressed CSI representation method; \textbf{AARES-CNN}~\cite{zhang2023csi}, an attention-augmented residual CNN; and \textbf{TIPS}~\cite{zhang2022tips}, a Transformer-based system with global self-attention.

As shown in Figure~\ref{fig:sota_comparison}, MambaLoc ($0.230\,\mathrm{m}$) outperforms DeepFi ($0.318\,\mathrm{m}$), AARES-CNN ($0.240\,\mathrm{m}$), and TIPS ($0.238\,\mathrm{m}$), demonstrating that a lightweight distilled student can match and exceed heavy Transformer-based models when guided by a privileged teacher.

\begin{table}[h]
 \centering
 \caption{Efficiency Comparison. MambaLoc achieves the best efficiency--accuracy trade-off.}
 \label{tab:efficiency}
 \begin{tabular}{|l|c|c|c|}
   \hline
   \textbf{Model} & \textbf{GFLOPs} & \textbf{Latency (ms)} & \textbf{Params} \\
   \hline
   TIPS (Transformer) & $1.12$ & $16.55$ & $447$k \\
   \hline
   AARES-CNN & $0.70$ & $12.0$ & $767$k \\
   \hline
   \textbf{MambaLoc (Ours)} & \textbf{0.39} & \textbf{6.94} & \textbf{257k} \\
   \hline
 \end{tabular}
\end{table}

\subsection{Computational Efficiency Analysis}

\subsubsection{Experimental Setup}
All experiments use PyTorch on a single NVIDIA GeForce GTX 1650 GPU (4\,GB VRAM), representative of mid-range edge computing hardware. Training uses AdamW with cosine annealing: the UWB teacher is trained for 40 epochs (lr $2{\times}10^{-3}$); the CSI student uses lr $5{\times}10^{-4}$ for stable convergence during knowledge distillation; batch size is 32. Latency is measured with CUDA synchronization; FLOPs computed via \texttt{thop}; all models benchmarked at $B{=}1$ for single-sample inference.

\subsubsection{Results}

Table~\ref{tab:efficiency} shows that MambaLoc achieves $6.94\,\mathrm{ms}$ inference---$2.38{\times}$ faster than TIPS ($16.55\,\mathrm{ms}$) and $1.73{\times}$ faster than AARES-CNN---a direct consequence of replacing quadratic self-attention with linear-time recurrent inference. MambaLoc requires only $0.391$\,GFLOPs per forward pass (65.1\% reduction vs.\ TIPS) and has the smallest footprint at $2.57{\times}10^5$ parameters, 42.5\% fewer than TIPS and $2.98{\times}$ fewer than AARES-CNN. On battery-powered edge devices where compute, memory, and power are primary constraints, these reductions translate to feasibility of continuous real-time localization without dedicated accelerator hardware.

\subsection{Ablation Studies}

\begin{figure}[h]
   \centering
   \includegraphics[width=1\linewidth]{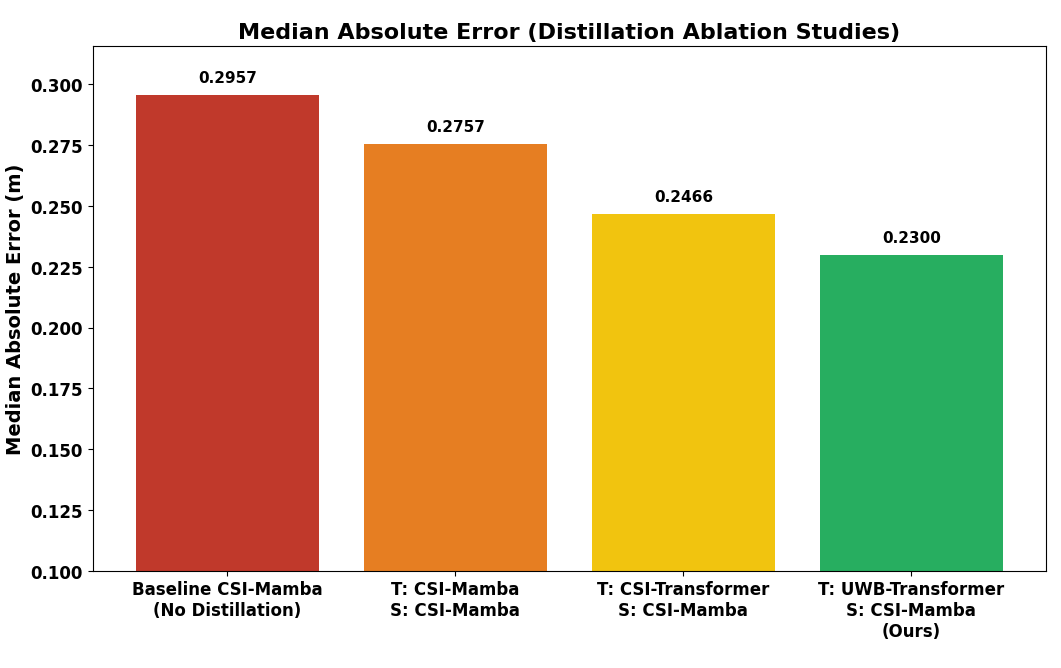}
   \caption{\textbf{Impact of Distillation Strategies.} Progressive improvements in localization accuracy (lower MAE is better) as distillation components are introduced.}
   \label{fig:ablation}
\end{figure}
\begin{figure*}[t]
 \centering
 \includegraphics[width=1\textwidth]{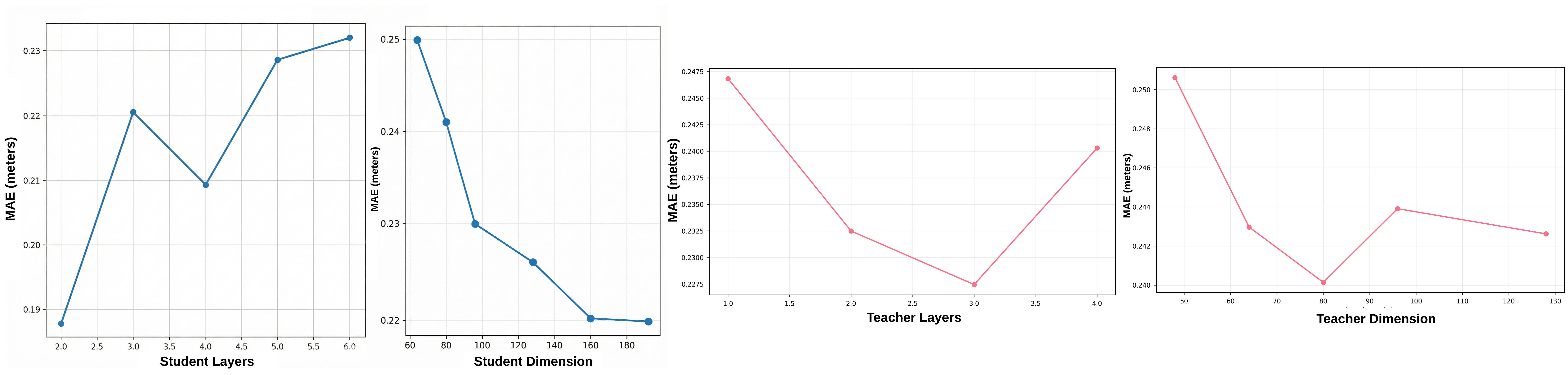}
 \caption{\textbf{(a,b)} The CSI student favors a ``shallow-wide'' design to capture high-dimensional CSI features without overfitting noise. \textbf{(c,d)} The UWB teacher exhibits a complexity ``sweet spot'' (3 layers, 80 dim); exceeding this widens the capacity gap, hindering effective distillation.}
 \label{fig:wide}
\end{figure*}
\vspace{-10pt}
\begin{figure}[h]
   \centering
   \includegraphics[width=0.7\linewidth]{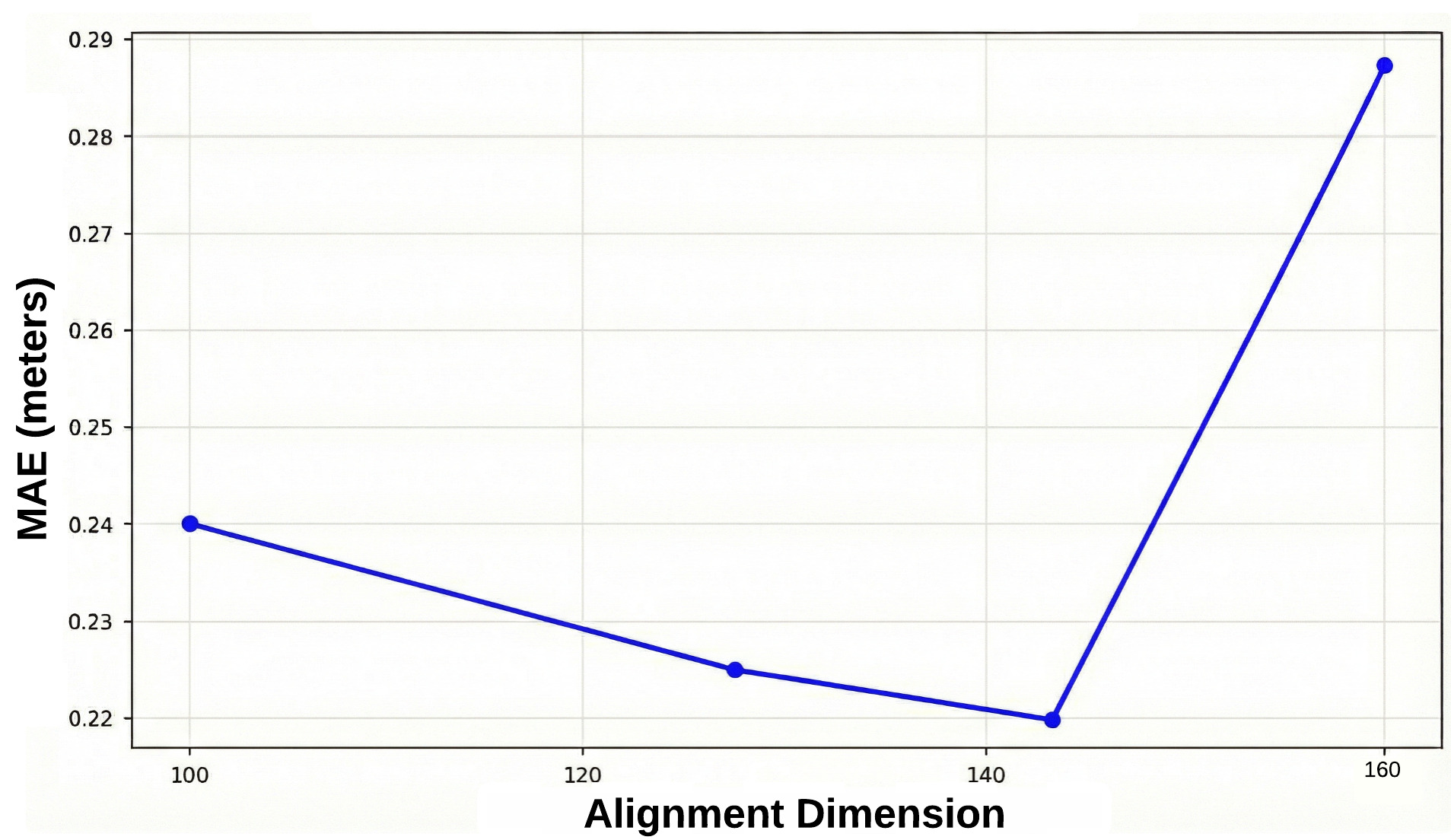}
   \caption{A dimension of 144 balances capacity with regularization; lower values cause bottlenecks ($<120$) and higher values permit overfitting ($>150$).}
   \label{fig:aligner}
\end{figure}
\begin{figure}[h]
   \centering
   \includegraphics[width=0.75\linewidth]{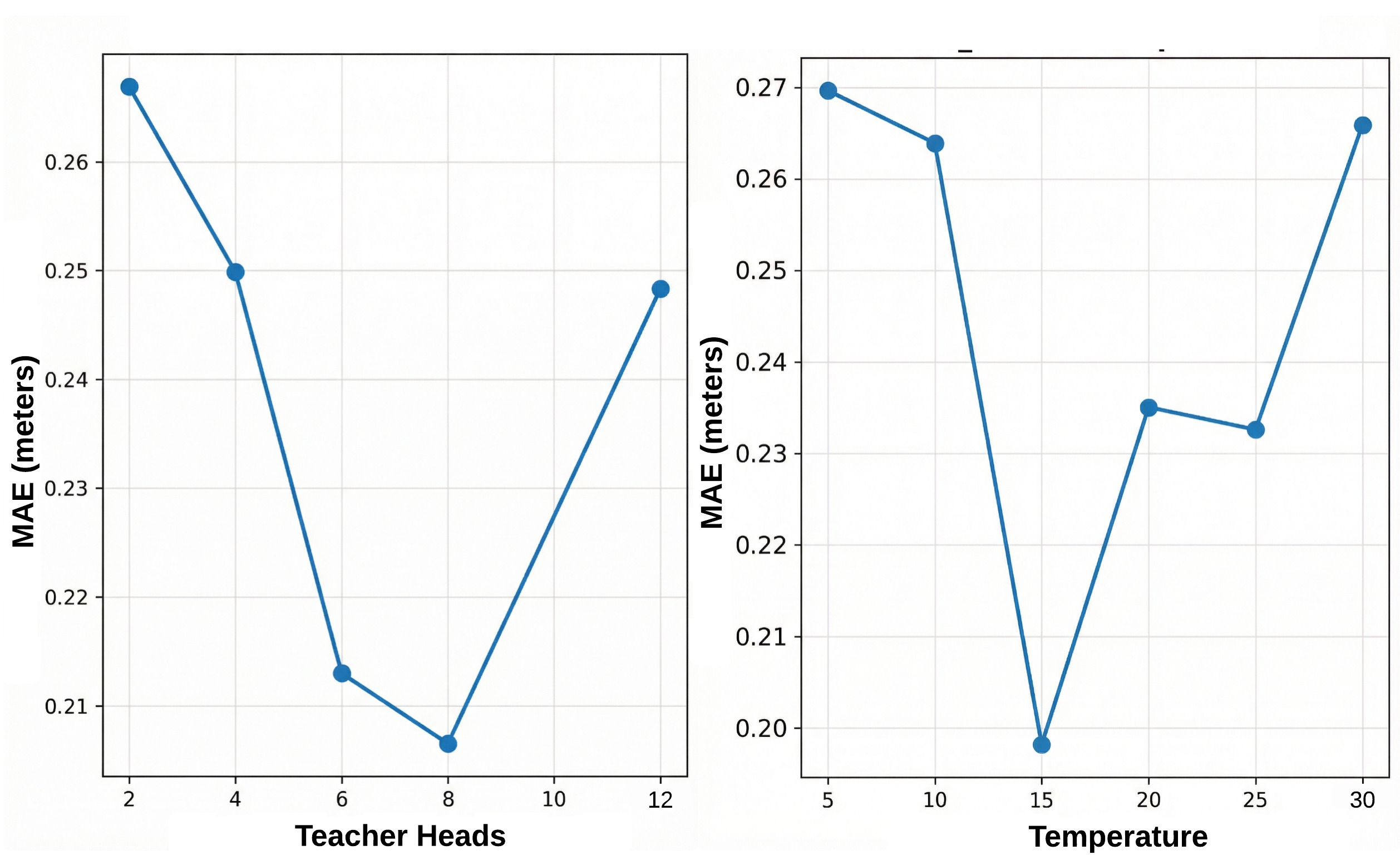}
   \caption{\textbf{(a)} Localization error is minimized at 8 attention heads. \textbf{(b)} A minimum at $T{=}15$ balances distribution sharpness with entropy.}
   \label{fig:1by2}
\end{figure}

\subsubsection{Distillation Strategy}
Figure~\ref{fig:ablation} summarizes incremental gains:
\begin{itemize}
   \item \textbf{CSI-Mamba (no teacher):} Median AE $= 0.2957\,\mathrm{m}$, serving as the lower bound.
   \item \textbf{Same-modality distillation} (CSI-Mamba $\rightarrow$ CSI-Mamba): $0.2757\,\mathrm{m}$, confirming distillation benefits while highlighting the limits of non-privileged teachers.
   \item \textbf{Cross-architecture distillation} (CSI-Transformer $\rightarrow$ CSI-Mamba): $0.2466\,\mathrm{m}$, demonstrating the value of distilling from attention-based architectures.
   \item \textbf{MambaLoc} (UWB-Transformer $\rightarrow$ CSI-Mamba): $0.2300\,\mathrm{m}$, proving UWB provides critical geometric cues inaccessible from CSI alone and validating the cross-modal, cross-architecture approach.
\end{itemize}
\subsubsection{Architectural Ablation}
Figure~\ref{fig:wide} illustrates depth/width sensitivity for both models.

\textbf{Student:} Increasing latent dimension $d_{\text{model}}$ from 60 to 160 monotonically reduces error; a wider latent space is essential to preserve the high-dimensional CSI frequency-diversity cues ($D_c{=}540$) without premature compression. Depth beyond 2 layers degrades performance, as deeper networks overfit CSI noise artifacts rather than signal structure; a ``shallow-wide'' architecture acts as an effective regularizer.

\textbf{Teacher:} An optimal point exists at 3 layers and 80 dimensions. Before this, the teacher cannot accurately model UWB geometry. Beyond it, an overly complex teacher learns highly UWB-specific representations that the CSI student cannot imitate across the modality gap, increasing distillation error.

\subsubsection{Hyperparameter Sensitivity}

\begin{itemize}
   \item \textbf{Alignment dimension} (Figure~\ref{fig:aligner}): Sharp optimum at $D{=}144$. Dimensions below 120 create information bottlenecks; above 150 the student maps noise to sparse dimensions, failing to learn meaningful geometry.

   \item \textbf{Attention heads} (Figure~\ref{fig:1by2}a): Error is minimized at 8 heads, likely corresponding to dominant multipath components (e.g., LOS and primary reflections). Increasing to 12 heads over-fragments the signal, reducing semantic coherence.

   \item \textbf{Distillation temperature} (Figure~\ref{fig:1by2}b): Static-temperature analysis reveals a ``V-curve'' with optimum at $T{=}15$; below 10 distributions are too sharp to capture spatial ambiguity, and above 20 they are too flat. These findings motivate our annealing schedule ($30{\to}10$): the high initial temperature prevents early commitment to specific spatial peaks, forcing global structure learning; the sweep through $T{\approx}15$ provides effective distillation; ending at $T{=}10$ acts as a precision refinement phase, unlike starting training with a static $T{=}10$, which causes failure, ending there after global structure is encoded achieves high-precision localization.
   \end{itemize}

\section{Conclusion}

We presented MambaLoc, a CSI-based localization framework combining linear-time SSMs with cross-modal distillation using UWB as privileged training information. Results demonstrate that the modality gap between UWB and CSI is reduced by aligning intermediate feature representations rather than relying on output-level distillation alone. The three-stage curriculum and adaptive weighting strategy are critical for stable convergence when distilling across sensing modalities and architectures, achieving a median localization error of 0.230\,m with 65.1\% fewer FLOPs and 58\% lower latency than Transformer-based approaches. Future work will evaluate the framework in cross-building scenarios~\cite{hesham2024localingua}  and investigate domain adaptation for cross-environment generalization. The distillation framework may also be extended to other RF sensing tasks such as activity recognition and environmental mapping, and to additional sensing modalities, directions that may clarify the role of cross-modal supervision and efficient sequence modeling in edge-native RF perception systems.

\bibliographystyle{IEEEtran}
\bibliography{references}

\end{document}